\documentclass[11pt]{article}
\usepackage[utf8]{inputenc}
\usepackage[T1]{fontenc}
\usepackage{graphicx}
\usepackage{grffile}
\usepackage{longtable}
\usepackage{wrapfig}
\usepackage{rotating}
\usepackage[normalem]{ulem}
\usepackage{amsmath}
\usepackage{textcomp}
\usepackage{amssymb}
\usepackage{capt-of}
\usepackage{hyperref}
\usepackage{color}
\usepackage{listings}
\usepackage[onehalfspacing]{setspace}
\usepackage[margin=1in]{geometry}
\usepackage{graphicx}
\usepackage{color}
\usepackage{multirow}
\usepackage{multicol}
\usepackage{amsfonts}
\usepackage{hyperref}
\usepackage{amsmath,amssymb}
\usepackage[english]{babel}
\usepackage{multimedia}
\usepackage{natbib}
\usepackage{subfig}
\usepackage[Procedure,boxed]{algorithm}
\usepackage[noend]{algpseudocode}
\author{ Oleg V.~Shylo  and Hesam Shams \\
Department of Industrial and Systems Engineering\\
University of Tennessee\\
\texttt{oshylo@utk.edu, hshams@vols.utk.edu} \\
}
\usepackage{fancyhdr}
\fancyheadoffset{0pt}
\date{\today}
\title{Boosting Binary Optimization via Binary Classification: A Case Study of Job Shop Scheduling}
\hypersetup{
 pdfauthor={Oleg Shylo, Hesam Shams},
 pdftitle={Boosting Binary Optimization via Binary Classification: A Case Study of Job Shop Scheduling},
 pdfkeywords={},
 pdfsubject={},
 pdfcreator={Emacs 25.2.2 (Org mode 9.1.13)}, 
 pdflang={English}}
\begin{document}

\maketitle
\begin{abstract}
Many optimization techniques evaluate solutions consecutively, where the next candidate for evaluation is determined by the results of previous evaluations. For example, these include iterative methods, ``black box'' optimization algorithms, simulated annealing, evolutionary algorithms and tabu search, to name a few. When solving an optimization problem, these algorithms evaluate a large number of solutions, which raises the following question: Is it possible to learn something about the optimum using these solutions? In this paper, we define this ``learning'' question in terms of a logistic regression model and explore its predictive accuracy computationally. The proposed model uses a collection of solutions to predict the components of the optimal solutions. To illustrate the utility of such predictions, we embed the logistic regression model into the tabu search algorithm for job shop scheduling problem. The resulting framework is simple to implement, yet provides a significant boost to the performance of the standard tabu search.

\bigskip

\noindent \emph{Key words:} binary optimization; machine learning; logistic regression; tabu search; job shop scheduling; probability dominance plots.
\end{abstract}
\noindent\hrulefill

\section{Introduction}
\label{sec:orgba31b8b}
\label{intro}

There is a class of algorithmic techniques that are employed to find the best option from a finite set of possibilities \(\mathcal{X}\). Many such algorithms scan solutions from \(\mathcal X\) successively and evaluate their quality according to some  objective function \(f: \mathcal X \rightarrow \mathbb{R}^1\).
In the process of solving an optimization problem, an algorithm produces a time-ordered sample of visited solutions and the corresponding objective values, \(\mathcal{S}=\{(x_1, f(x_1)), \ldots, (x_m, f(x_m): x_i \in \mathcal X, i=\overline{1,m}\}\). Is there anything to learn from these samples about the properties of the optimal solutions? For example, what is the probability that the component \(j\) is equal to one in some optimal solution, given the specific time-ordered sample. 

Clearly, the answer to this question depends on the algorithm that produced the sample, as well as on the particular problem instance. The samples produced by a random search may not be as useful as those generated by a more advanced algorithm. Furthermore, larger data samples can provide better information to support statistical inferences. 

We take a machine learning approach to the above questions and formulate the ``learning problem'' as a binary classification problem in Section \ref{ml.bin}. We apply logistic regression and explore the quality of the model predictions on a set of scheduling benchmarks, utilizing data samples generated by the tabu search \citep{Glover-1997-tabu-searc}. Section \ref{gta} demonstrates how to make use of the solution samples to boost the performance of the tabu search. We conclude in Section \ref{conclusions} with the potential extensions of this research.

The contributions presented in this paper are as follows. 
\begin{itemize}
\item The paper describes a general statistical learning model for binary optimization and establishes an efficient implementation based on logistic regression.
\item We empirically explore the computational performance of the logistic regression model on the job shop scheduling problem instances.
\item We provide a simple extension of tabu search that includes the proposed learning component. We demonstrate that such an extension is easy to implement, yet can provide a significant boost to the computational performance.
\item A novel framework for comparing optimization algorithms based on the concept of probability dominance is introduced, which provides intuitive and statistically sound tools for the experimental analysis of optimization algorithms.
\end{itemize}

\section{Literature Review}
\label{sec:org3951365}

The machine learning community has produced a tremendous amount of successful applications in recent years, ranging from image recognition to autonomous driving and outplaying humans in complex games. 
This spree of newsworthy results renewed the interest in combining the perspectives of artificial intelligence and operations research. Interestingly, many of the recent publications trace the ideas summarized by \citet{glover1986future} more than 30 years ago. In particular, the paper outlines a ``learning experiment'', which consists of an extensive data collection on a given class of problems, followed by a search for hidden patterns to enhance the solution process. The earliest application of this idea to the job shop scheduling problem was explored by \citet{Crowston-1964-prob-param}, where the authors conduct a learning experiment to derive probabilistic and parametric scheduling rules. The  contributions that followed can be roughly partitioned into three groups. 

The first group uses learning experiments to tune optimization algorithms, which can be very sensitive to the choice of internal parameters. Such calibration requires computationally expensive experimentation, and many approaches automate this process using controlled experimentation \citep{hutter2009paramils,Adenso-Diaz-2006-fine-tunin,Birattari-2002-racing-algo}.

The second group of publications considers the problem of matching problem instances to algorithms and their parameters. These are rooted in the idea that some algorithms perform well on certain instances but not others, so the goal is to exploit such situations by choosing from the set of available options. For example, one might construct a portfolio of available algorithms and make use of a large representative set of instances to build a prediction model that links problem features to the performance of available algorithms. These predictions are used to choose an algorithm from the portfolio on a per-instance basis, in order to achieve efficient performance on average when solving a sequence of optimization instances  \citep{Xu-2008-satzil,Malitsky-2013-algor-portf,Streeter-2012-new-techn}. A general overview of the algorithm selection as well as a benchmark library for algorithm selection is discussed by \citet{bischl2016aslib}. Other researchers took this idea even further and use selection procedures to adjust parameters and subroutines dynamically during the search process. In the context of branch and bound algorithms, the dynamic selection of branching heuristics was explored by \citet{di2016dash}. \citet{nannicini2011probing} used a support vector machine (SVM) classifier to guess whether the probing algorithm should be applied on a given node of the branch-and-bound tree.

The third group of publications, which is the closest to the ideas proposed in this paper, uses learning experiments to synthesize new optimization procedures. For example, \citep{alvarez2017machine} used  supervised learning to design a new branching strategy that is trained to imitate strong branching. Similarly,  \citet{khalil2016learning} outlined procedures that predict strong branching scores to accelerate the branching process. 
A node selection procedure was derived by \citet{he2014learning} using machine learning. A comprehensive overview of the machine learning applications for mixed integer programming solvers was performed by  \citet{lodi2017learning}. Furthermore, \citet{dilkina2017comments} and \citet{louveaux2017comments} made comments on the survey and provided some future directions for intelligent branch-and-bound for mixed integer programming. \citet{kruber2017learning}  developed a supervised learning approach to choose the best possible reformulation for decomposition. The experiments on structured instances showed a meaningful improvement on performance. \citet{basso2017random} used an unsupervised learning model to identify strong decomposition candidates. In the context of deep learning, the design of an optimization algorithm was defined as a learning problem by \citet{andrychowicz2016learning}, where an algorithm learns the structure of optimization problems.  Through learning, the algorithm automatically tunes its parameters to optimize the performance on a particular class of optimization problems. 

\section{Binary Classification for Binary Optimization}
\label{sec:orgc4c9526}
\label{ml.bin}

A binary optimization problem can be formally expressed as

\begin{equation}\label{general.binary.model}
\begin{array}{cc}
\text{minimize } f(x) \\
\text{s.t. } x \in \mathcal X \subset \{0,1\}^n,
\end{array}
\end{equation}
where the objective function \(f\) is an arbitrary function, and \(\{0,1\}^n \cap \mathcal X \neq \emptyset\).

Since each component of a feasible vector \(x\) is either 0 or 1, the index set \(\overline{1,n} \equiv \{1,\ldots,n\}\) can be partitioned into two subsets based on an optimal solution \(x^*\) to the problem (\ref{general.binary.model}). The first subset includes all indexes \(j\) for which \(x^*_j\) is equal to one, while the remaining indexes are assigned to the second subset. Formally, these classes are denoted \(\mathcal C^1=\{j | x_j^*=1\}\) and \(\mathcal C^0=\{j | x_j^*=0\}\), \(j\in \overline{1,n}\). 
Note that there are as many such partitions as there are optimal solutions.

\looseness=-1 Here we focus on the iterative methods, which transition  from one solution to another, generate a sequence of binary solutions, and evaluate the corresponding objective function values.  Consider a set of such solutions, \(\{x^1, \ldots, x^m\}\subseteq \mathcal X\), discovered after time \(t\). For \(j\in \overline{1,n}\), let \(I_j(t)\) denote a time-ordered list of pairs formed by the \(j\text{-th}\) components of these solutions and the corresponding objective values:

\begin{equation*}
I_j(t)=[ (x^1_j, f(x^1)), (x^2_j,f(x^2)), \ldots, (x^m_j, f(x^m))].
\end{equation*}

Assuming that every run of the algorithm produces a different vector \(I_j(t)\), we would like to classify \(j\) either as belonging to \(\mathcal C_1\) or \(\mathcal C_0\). In other words, we would like to build a prediction model that would map vectors \(I_j(t)\), \(j \in \overline{1,n}\), into the interval [0,1], estimating a conditional probability \(P(j\in \mathcal C^1 | I_j(t))\). If there are no consistent patterns in \(I_j(t)\), the model should return probabilities close to 0.5, otherwise the probabilities should get closer to the ends of the interval [0,1], yielding predictions about the optimal solution components. Clearly, the model depends on a particular algorithm that produced the vector \(I_j(t)\), as well as on the time \(t\) used to collect the data sample. 

Consider the decomposition of \(I_j(t)\) into two sequences, \(I_j^1(t)\) and \(I_j^0(t)\), where
\begin{align*} 
I_j^1(t)=\{ (x^k_j, f(x^k)): (x^k_j, f(x^k))\in I_j(t), x^k_j=1\},\\
I_j^0(t)=\{ (x^k_j, f(x^k)): (x^k_j, f(x^k))\in I_j(t), x^k_j=0\}.
\end{align*}

Here, \(I_j^1(t)\) describes all visited solutions with the component \(j\) equal to one, and \(I_j^0(t)\) corresponds to the objectives of solutions with the component \(j\) equal to zero. One of the sequences may be larger than the other, which makes it difficult to parametrize the model. To avoid this, one can downsample (remove) the entries in the larger sequence to match the size of the smaller sequence. Any data point in the data set for our binary classification problem can be described as 

\begin{center}
\begin{tabular}{|cc|c|}
\hline
Input &  & Output\\
\hline
\(I_j^1(t)\) & \(I_j^0(t)\) & x\(^{\text{*}}_{\text{j}}\)\\
\hline
\end{tabular}
\end{center}

Hence, we arrived at a binary classification problem, which can be solved by a variety of statistical methods (e.g., logistic regression, decision trees, random forests, neural networks). In the following discussion, we focus on the classification based on the logistic regression method. 

Logistic regression is a special case of a generalized linear model \citep{Agresti-2002-logis-reg}, and it is one of the most widely used statistical techniques for discrete data analysis. It can describe a relationship between one dependent binary variable and one or more nominal, ordinal, or interval independent variables. In other words, for a binary output variable \(Y\) and an input vector \(X\), the model  predicts the conditional probability \(p(Y=1|X) \equiv h_{\theta}(X)\). The hypothesis function \(h_{\theta}\) in the logistic regression is given by the sigmoid function
\begin{equation}\label{hypo.logreg}
h_{\theta}(X) = \frac{1}{1+e^{-(\theta_0 + \theta X)}}
\end{equation}

When fitting (\ref{hypo.logreg}) to a particular dataset  \(\{X^i,Y^i\}\), the parameter \(\theta\) can be obtained via the maximum likelihood method \citep{Kleinbaum-2010-maxim-likel}:
\begin{equation*}
\theta^* = \arg \max \prod_{i}  h_{\theta}(X^i)^{Y^i} (1-h_{\theta}(X^i))^{1-Y^i}
\end{equation*}

Now, the logistic regression hypothesis for our binary classification problem \(h_{\theta(t)}( I_j(t) )\) can be stated as 
\begin{equation} \label{HF.full}
h_{\theta(t)}(  I_j(t) ) =\frac{1}{1+\exp{\left[\displaystyle \sum_{ (1,f(x_k))\in I_j^1(t) } \theta(t)f(x_k)-\displaystyle \sum_{ (0,f(x_k))\in I_j^0(t) } \theta(t) f(x_k)\right]}}  
\end{equation}

Note that there is a single parameter in this model, \(\theta(t)\in \mathbb{R}^1\). There are multiple reasons for this restriction. First,  assume that each variable \(x_j\) in (\ref{general.binary.model}) is substituted with \(x^{new}_j=\mathbf{1}-x_j\). Clearly, this substitution would not change the optimal objective. However, the terms in  (\ref{HF.full}) would switch from one sum to another. Hence, by using the same parameter \(\theta(t)\) for the left summation and for the right summation, we guarantee that the equivalent encodings would produce the same model. Similarly, if there was a different parameter \(\theta(t)\) for each of the terms in summations, the order in which the solutions were visited would matter. However, we would like to avoid this temporal dependence by using a single parameter \(\theta(t)\). Lastly, all vectors \(I_j(t)\) are treated similarly to avoid reliance  on the naming conventions, so the index \(j\) in \(I_j(t)\) does not play any role in predictions. As a result, the data for different solution components can be merged by simply concatenating the information vectors, \(I_j(t)\), \(j\in \overline{1,n}\), and their corresponding labels, \(\mathcal C_1\) or \(\mathcal C_0\), into a single data set. 

The size of \(I_j(t)\) changes with every new solution, and the optimal value of \(\theta(t)\) may change for different sizes of \(I_j(t)\). To avoid this, we can project \(I_j(t)\) to a smaller, fixed dimension. For example, such reduction can be achieved using the minimum function. Denote the
resulting values by \(D^1_j(t)\) and \(D^0_j(t)\):
\begin{align} \label{D1D0}
\begin{split}
D^1_j(t)&=\min( \{ f(x_j^k) | (x_j^k, f(x^k)) \in I_j(t), x_j^k=1\}) \\
D^0_j(t)&=\min( \{ f(x_j^k) | (x_j^k, f(x^k)) \in I_j(t), x_j^k=0\}) 
\end{split}
\end{align}

This provides us with the reduced information vector \(I^R_j(t)=[D^1_j(t), D^0_j(t)]\in \mathbb R^2\). Clearly, instead of storing \(I_j(t)\) in the memory for these calculations, \(I_j^R(t)\) can be updated directly every time the algorithm finds a new feasible solution. 

The corresponding reduced logistic regression model is described using the hypothesis 

\begin{equation}\label{HF.compact}
h_{\theta(t)}(  D^1_j(t) , D^0_j(t) ) =\frac{1}{1+\exp{\left(\theta(t) [D^1_j(t) -  D^0_j(t)]\right)}} 
\end{equation}

\looseness=-1 To clarify the classification problem, consider an example of the training data in Table \ref{table.logreg}. Two independent variables, \(D^1(t)\) and \(D^0(t)\) are used to predict whether the corresponding component in an optimal solution is equal to zero or one, with the ground truth values in column \texttt{opt}. To collect the training data for columns \(D^1(t)\) and \(D^0(t)\), the algorithm runs repeatedly for time \(t\) and reports the best objective for each solution component. Note that an algorithm may never encounter certain solution components with a value of zero (or one). In this case, we omit the corresponding rows from the data set, and the model prediction is fixed to 0.5 for those components.  

\begin{table}[htbp]
\caption{An example of the training data for the logistic regression model. \label{table.logreg}}
\centering
\begin{tabular}{|c|c|c|}
\hline
D\(^{\text{1}}\)(t) & D\(^{\text{0}}\)(t) & opt\\
\hline
1395 & 1366 & 0\\
1368 & 1400 & 1\\
1366 & 1438 & 1\\
1373 & 1366 & 0\\
1379 & 1365 & 0\\
1365 & 1389 & 1\\
\hline
\end{tabular}
\end{table}

\begin{figure}[!ht] \centering
\includegraphics[width=0.8\textwidth]{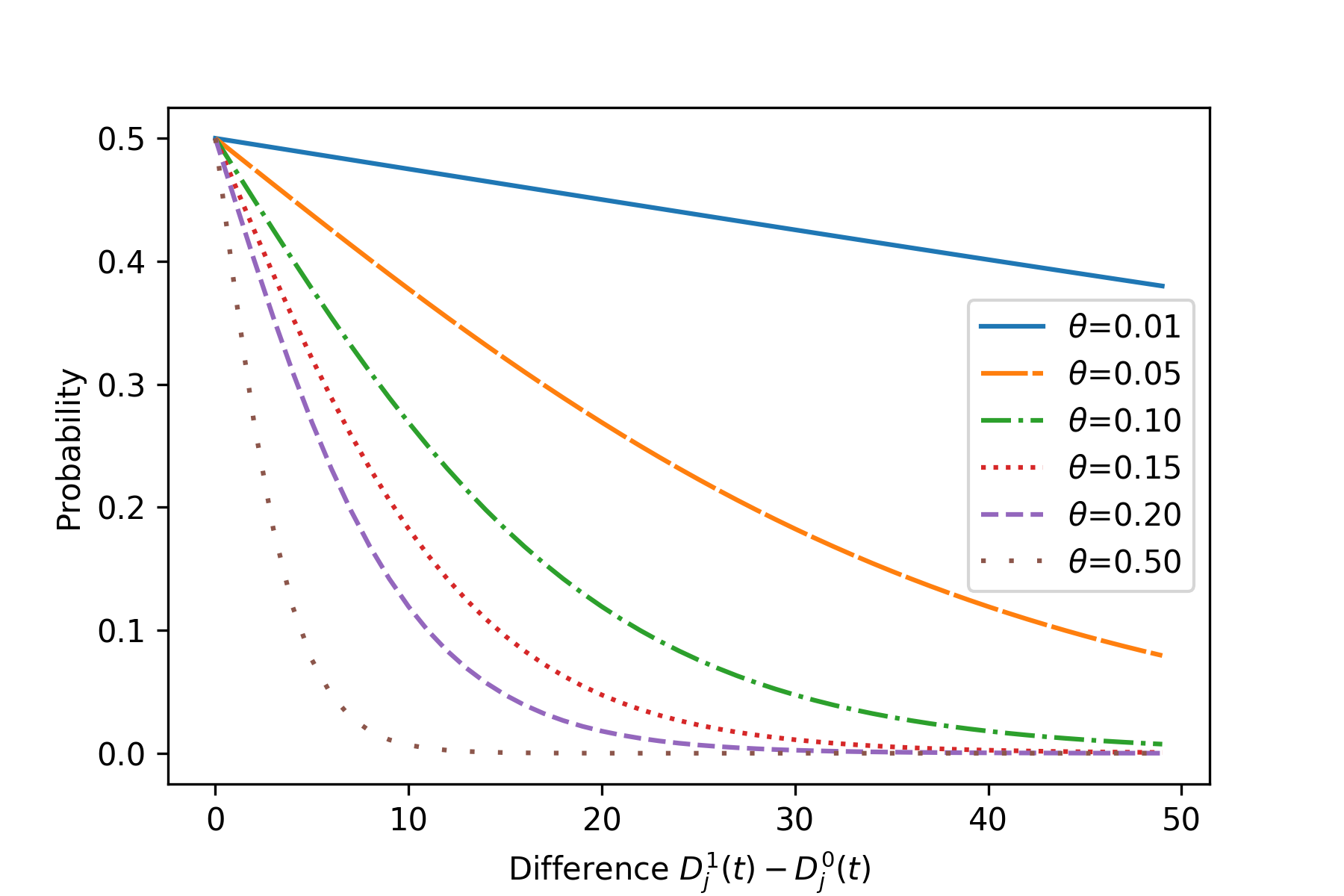}
     \caption{Logistic regression predictions for different values of the regression parameter $\theta$.}
     \label{fig:regression.example}
   \end{figure}

The model predictions for the component \(j\) are solely based on the difference between the best objective found with \(x_j=1\) and the best found objective with \(x_j=0\). Figure \ref{fig:regression.example} shows the predicted probabilities as a function of this difference.

In practice, the model (\ref{HF.compact}) can be tuned for a representative set of optimization problems, for which optimal or high-quality reference solutions are already established. Then the tuned model can be applied to new problem instances without known solutions, assuming that the new instances are ``similar'' to the problem set used to parametrize the model. 

The hypothesis functions (\ref{HF.full}) and (\ref{HF.compact}) first appeared in the context of the Global Equilibrium Search \citep{Shilo:1999}, which has shown excellent performance on a variety of classic combinatorial problems: job shop scheduling \citep{Pardalos:2006}, weighted maximum satisfiability \citep{Shylo:2008a}, quadratic unconstrained binary optimization \citep{Pardalos:2008}, and the maximum cut problem \citep{Shylo:2010,Shylo:2011a}, to name a few. Interestingly, these formulas were derived by \citet{Shilo:1999} by assuming the Boltzmann's distribution over the set of all feasible solutions, without any reference to the logistic regression or maximum likelihood estimation. 

In the next section, we explore the empirical distribution of the optimal regression parameter \(\theta(t)\) for different values of the time threshold \(t\) for a set of job shop scheduling instances \citep{Jain:1999}, and apply an iterative  algorithm based on tabu search \citep{Nowicki:1996} to collect the training data. The choice of the tabu search is motivated by the wide domain of successful applications in the literature, while the job shop scheduling problem captures many complexities common to binary optimization domains.

\subsection{Computational Experiment: Logistic Regression for Binary Optimization}
\label{sec:orgffcc0d7}
\label{log.comp.sec}

Note that the optimal regression parameter \(\theta(t)\) in (\ref{HF.compact}) depends on a specific training dataset, which in our case, is generated by an optimization algorithm solving a problem instance for a fixed number of iterations. To collect the data, we need to choose an optimization algorithm, specific problem instances to solve, and a time threshold for data collection. The proposed framework was tested using the tabu search algorithm \citep{Glover-1997-tabu-searc} applied to the job shop scheduling problem. As a general purpose technique, the tabu method does not use any problem specific features of the job shop scheduling formulation and mainly relies on exploring the search space through a sequence of local moves. Hence, the following discussion should be viewed as a proof of concept, and is neither limited  to the job shop scheduling problem, nor to the tabu search. The same methodology can be applied to any iterative binary optimization algorithm. 

\subsubsection{Problem Instances}
\label{sec:org978feae}

In the job shop scheduling problem, there is a set \(M=\{M_1,M_2,\ldots,M_m\}\) of machines and a set \(O=\{\sigma_1,\ldots,\sigma_{N}\}\) of operations. To complete operation \(\sigma_i\in O\), it has to be processed by the specific machine \(M(\sigma_i)\in M\) for time \(p(\sigma_i)\). Additionally, there are precedence constraints between operations defined by a set \(\mathcal{P}\) of ordered pairs, so if \((\sigma_i,\sigma_j)\in \mathcal{P}\), then \(\sigma_i\) has to be finished before starting \(\sigma_j\). If two operations are linked by the precedence constraint, we say that they belong to the same job. The objective is to find a processing order of operations to minimize the largest completion time. 

For any two operations \(\sigma_i\) and \(\sigma_j\) that are processed by the same machine, either \(\sigma_i\) or \(\sigma_j\) is processed first. Let \(x_{(i,j)}\) denote a binary variable that determines the order between \(\sigma_i\) and \(\sigma_j\). If there are \(n\) operations assigned to the same machine, we would need \(n\cdot(n-1)/2\) binary variables to define their processing sequence. Let \(I\) denote a set of all binary variable indexes, \(I=\{ (i,j): \sigma_i,\sigma_j \in O, i<j, M(\sigma_i)=M(\sigma_j)\}\), and let \(x\) denote a binary vector comprised of all \(x_{(i,j)}\) variables. Then the earliest completion time for the schedules defined by \(x\) can be found by solving the following problem.

\begin{align}
\begin{split}
 \text{minimize } C_{max}(x) := \max\limits_{\sigma_i \in O}[s(\sigma_i)+p(\sigma_i)] & \label{obj}\\ 
\text{subject to }& \\
s(\sigma_i)\geq 0 \mbox{, } \forall \sigma_i\in O &\\
s(\sigma_j)\geq s(\sigma_i)+p(\sigma_i) \mbox{, } \forall (\sigma_i,\sigma_j)\in \mathcal{P}&\\
s(\sigma_i)\geq s(\sigma_j) + p(\sigma_j) - C\cdot x_{(i,j)}, \forall (i,j)\in I\\
s(\sigma_j)\geq s(\sigma_i) + p(\sigma_i) - C\cdot (1-x_{(i,j)}), \forall (i,j)\in I. 
\end{split}
\end{align}

Here \(s(\sigma_i)\) provides the start time of the operation \(\sigma_i\), and \(C_{max}\) is often referred to as the 
makespan of the schedule represented by start times of operations from \(O\). The constant \(C\) is a large positive number, which can be set to \(\sum_{\sigma_i\in O} p(\sigma_i)\).
If the problem is infeasible, the optimal objective is considered to be equal to infinity. Let \(C^*_{max}(x)\) denote the optimal objective of the problem (\ref{obj}), then the job shop scheduling problem can be defined as 
\begin{align*}
\text{minimize } C^*_{max}(x) &\\
x_{(i,j)}\in \{0,1\}, & \forall (i,j)\in I. 
\end{align*}

For the computational experiment we considered Taillard's benchmark problems. This problem set contains 80 randomly generated problems denoted by ta01-ta80 generated by   \citet{Taillard:1993}. Optimal solutions are known for 58 problems from this class. In the literature the problems ta01-ta10 and ta51-ta80 are often reported as the easiest problems of the set, therefore we limited our testing to the remaining problems ta11-ta50. The ratio of jobs to machines in ta11-ta50 ranges from 1 to 2, and, as of now, no optimality proofs are known for instances with the ratio of 1.5.
A constraint programming search algorithm developed by \citet{Vilim-2015-fail-direc} is among the few successful applications of generic search algorithms to these instances. It is also an integral part of the IBM ILOG CP Optimizer. The study of commercial MIP solvers (Gurobi, CPLEX and SCIP) by \citet{Ku-2016-mixed-integ} revealed that the problems ta11-ta50 are still out of their reach.

\subsubsection{Optimization Algorithm}
\label{sec:orgf65c4cc}
\label{sec.tabu.search}

Tabu search proposed by \citet{Glover:1989} is arguably one of the best standalone optimization approaches among those based on local search. Tabu search employs a short-term memory prohibition mechanism that prevents the revisiting of solution attributes that have been recently removed. Less commonly, tabu restrictions inhibit the removal of attributes that were recently introduced. In general, these two types of restrictions lead to different search trajectories and might be employed in parallel, however in the case of 0-1 optimization problems, they are equivalent. Through inhibition mechanisms and by enabling non-improving solution attributes, the tabu search method provides an almost effortless escape from local minima together with efficient exploration of the search space. Notably, many of the best algorithms for the job shop scheduling utilized a tabu search component \citep{Zhang-2008-very-fast,Goncalves-2013-exten-akers,Peng-2015-tabu-searc}.

\begin{algorithm} \caption{Pseudocode of the tabu search algorithm. } \label{algo.tabu}
\begin{algorithmic}[1]
\Procedure{TabuAlgorithm}{$T_{min}$, $T_{max}$, $nepochs$, $niters$}
    \State generate a random solution $x^{min}$ 
    \State $x^0 = x^{min}$ \label{tabu.line.revert}
    \State $tabuExp(j)=0$, $j=1,\ldots,|x^{min}|$;   \Comment{initialize tabu status expiration}  
    \For{$epoch=1$ to $nepochs$}      
      \For{$iter=1$ to $niters$}           
           \State $tenure \leftarrow$ generate a random integer from $[T_{min},T_{max}]$ 
           \For{$x$ in $N(x^0)$} \label{tabu.line.scan1}
               \State $expir(x) = iter$ \label{tabu.line.status1}
               \For{$j$ in $\{j: x^0_j\neq x_j\}$}
                    \State $expir(x) = \max(tabuExp(j), expir(x))$                      
                \EndFor 
                \If{$expir(x)>iter$}
                   \State TabuSet = TabuSet $\cup \ x$
                \EndIf \label{tabu.line.status2}
            \EndFor            \label{tabu.line.scan2}
            \State NonTabuSet = $N(x^0)$ - TabuSet
            \If{NonTabuSet $\neq \emptyset$}
               \State $x^{new}= \arg \min \{f(x): x\in \text{NonTabuSet}\}$  \Comment{best non-tabu solution}
            \Else
               \State $x^{new} = \arg \min\{ expir(x): x\in N(x^0)\}$ \Comment{ tabu solution with earliest expiration}
            \EndIf        
            \For{$j$ in $\{j: x^{new}_j\neq x^0_j\}$} \Comment{only look at the components that have changed}
               \State $tabuExp(j)=iter+tenure$
            \EndFor            
            \State $x^0=x^{new}$
            \If{[$f(x^0)<f(x^{min})$]} 
               \State $x^{min}=x^0$
            \EndIf 
    \EndFor
    \EndFor
    \Return $x^{min}$
\EndProcedure
\end{algorithmic}    
\end{algorithm}

Procedure \ref{algo.tabu} describes our implementation of the tabu search method. The vector \(tabuExp\) keeps track of the tabu status of solution components. For example, if \(tabuExp[5]=100\), then the fifth component of the solution vector is considered tabu, as long as the iteration counter is less than 100. If there are multiple solution components involved, the tabu status is defined by the component with the largest value of \(tabuExp\). The algorithm runs for a fixed number of epochs, each consisting of \(niters\) iterations. In each iteration the duration of the tabu tenure is set to a random integer in the interval \([T_{min},T_{max}]\) to prevent cycling. The algorithm scans through all the solutions in the neighborhood of the current solution \(x^0\) (Procedure \ref{algo.tabu}, lines \ref{tabu.line.scan1}-\ref{tabu.line.scan2}) to identify the set of prohibited/tabu solutions. The non-tabu solution with the best objective value is chosen as the next incumbent solution, using  random tie-braking if necessary. If all of the solutions in the neighborhood are prohibited, then the algorithm chooses the solution with the earliest expiration of its tabu status. The best found solution is returned after the fixed number of search epochs.

Note that there only two components of the algorithm that are problem specific: the neighborhood \(N(x)\) and the specific objective function \(f(x)\). In our case,  the  search is based on the N4-neighborhood for the job shop scheduling problem \citep{Blazewicz:1996}, which makes use of the critical path concept. The N4-neighborhood consists of solutions obtained by moving operations either to the beginning or to the end of their critical blocks. As for the objective, we minimize the makespan of the schedule as defined in (\ref{obj}).   

\subsubsection{Computational Results}
\label{sec:org6a5a515}

In the first set of experiments, we investigated the predictive potential of the logistic regression model (\ref{HF.compact}). The tabu search outlined in the previous section was used to solve Taillard's instances ta11-ta50. In total, we performed 20 runs, each run was limited to \(nepochs=200\) and each epoch consisted of \(niters=300,000\) iterations. The lower tenure range parameter \(T_{min}\) was set to 5 and \(T_{max}\) was set to 11.

\begin{figure}[!ht]
     \subfloat{%
       \includegraphics[width=0.5\textwidth]{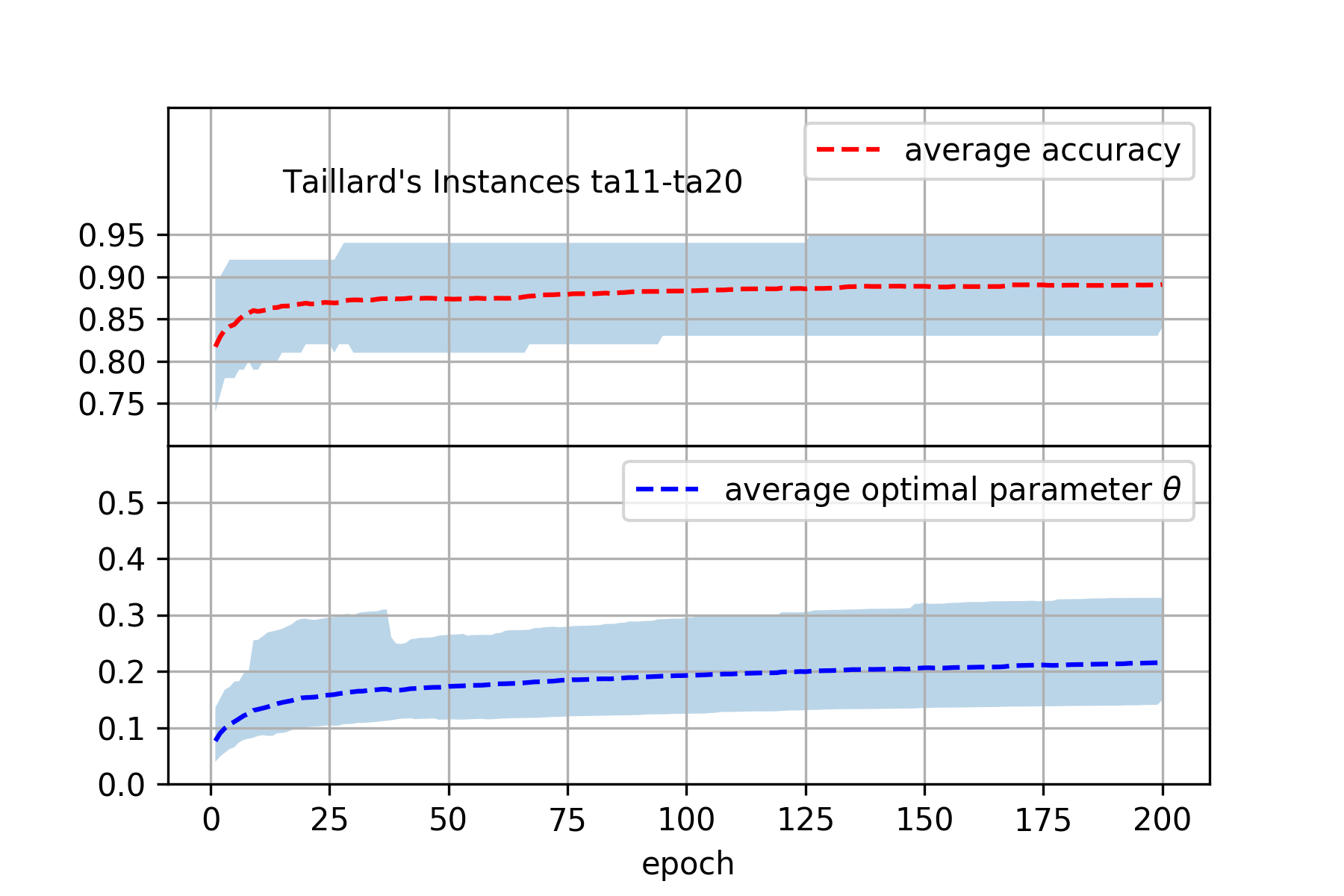}
     }
     \hfill
     \subfloat{%
       \includegraphics[width=0.5\textwidth]{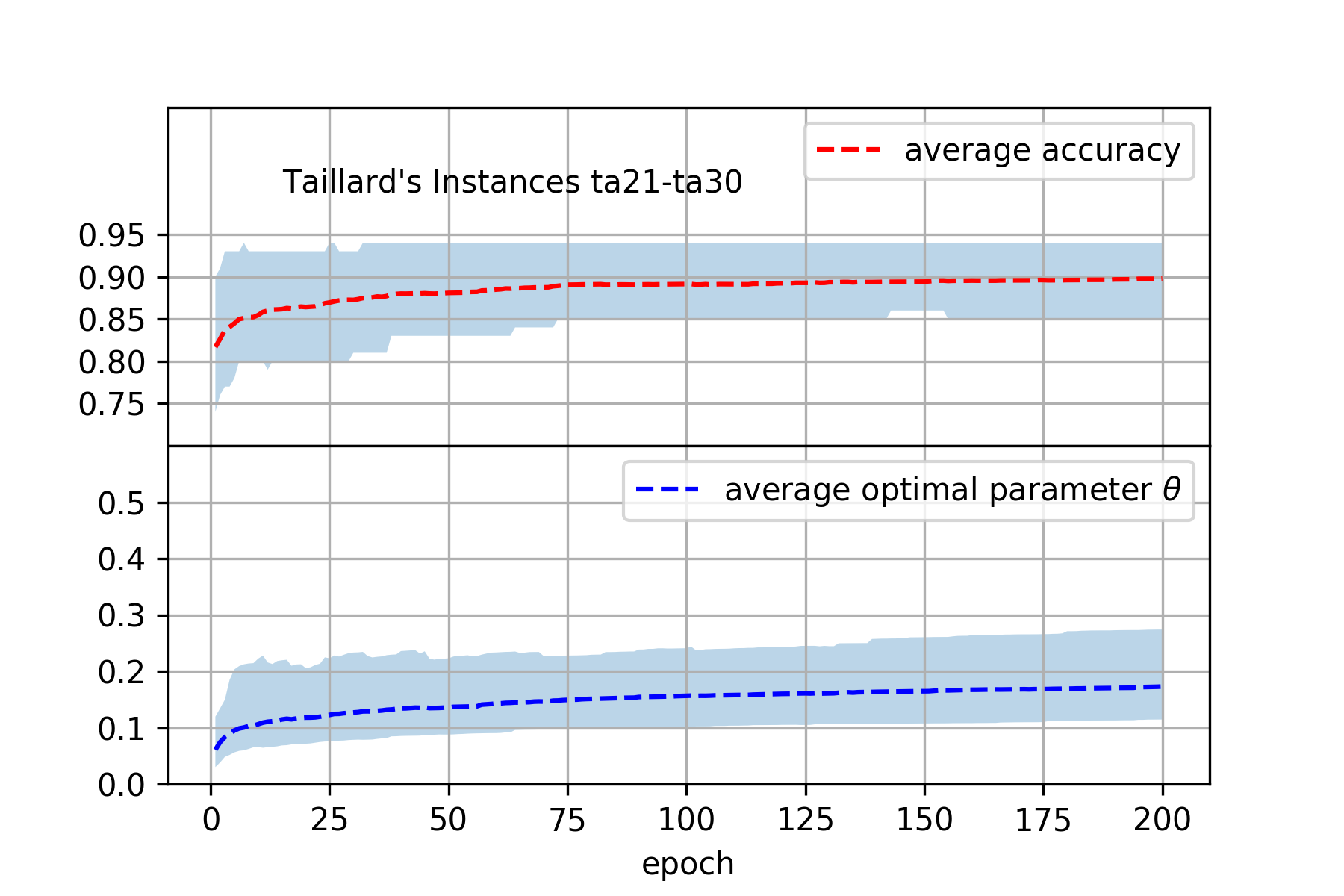}
     }
     \vfill
     \subfloat{%
       \includegraphics[width=0.5\textwidth]{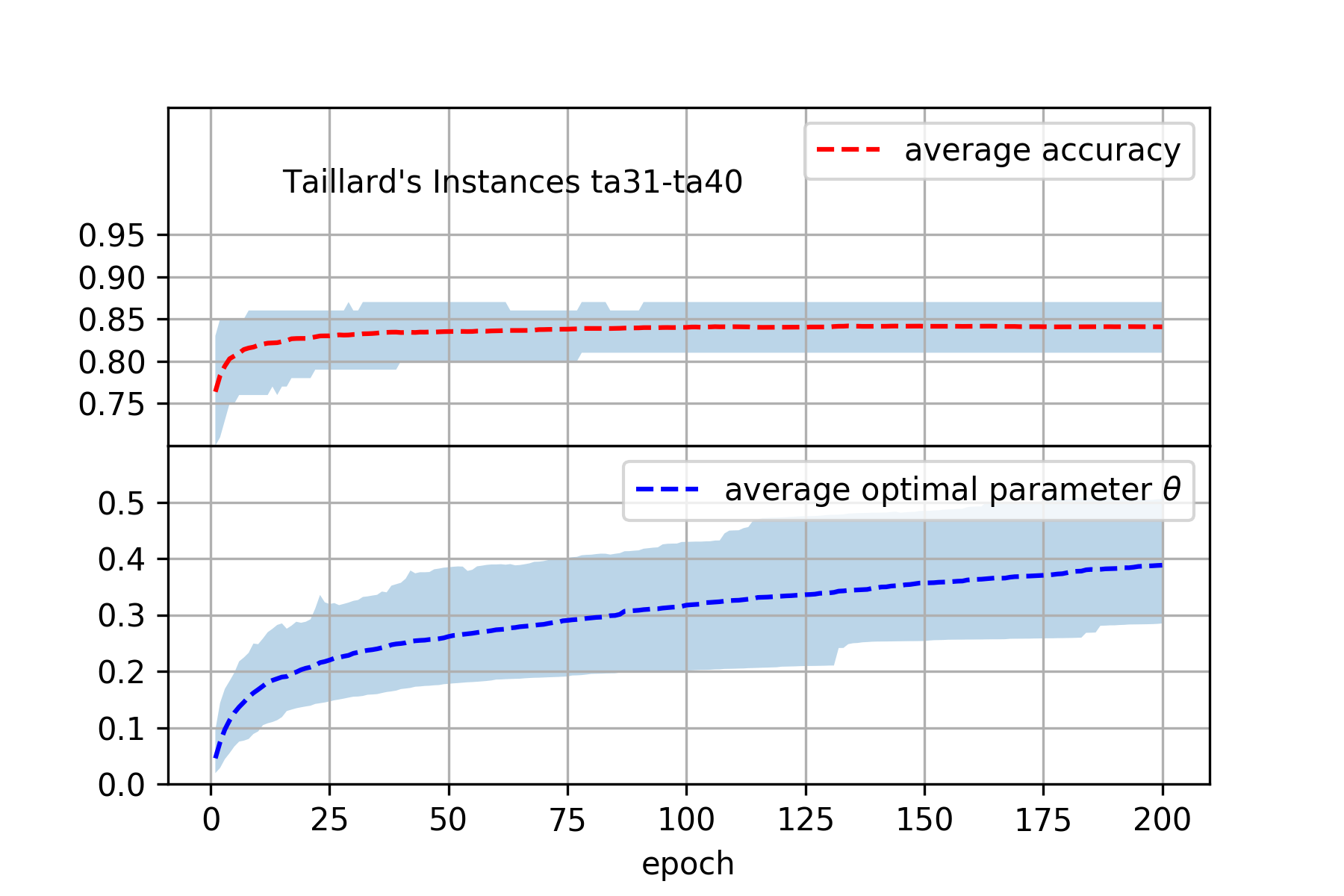}
     }
     \hfill
     \subfloat{%
       \includegraphics[width=0.5\textwidth]{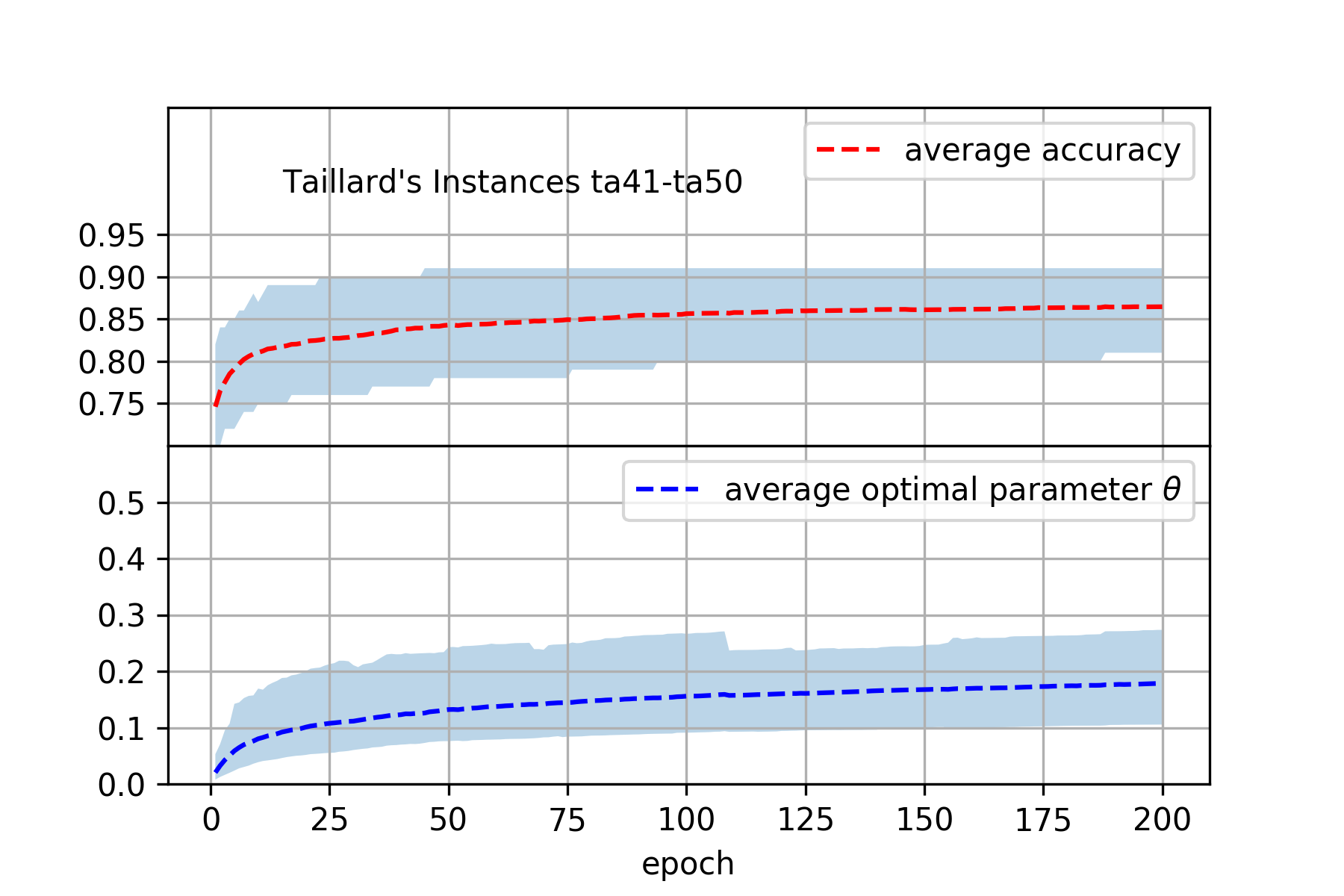}
     }
     \caption{Optimal regression parameter $\theta$ and the corresponding accuracy of the logistic regression model on the set of Taillard's instances ta11-ta50.}
     \label{fig:regression}
   \end{figure}

\looseness=-1 In every run we continuously updated the values \(D^1_j\) and \(D^0_j\) for each solution component \(j\) as defined in (\ref{D1D0}). Remember that these values are the best objectives found when the component \(j\) is equal to one and zero respectively, so each update to the vectors \(D^1_j\) and \(D^0_j\) takes \(O(|x|)\) steps, where \(|x|\) is the size of the binary solution vector. In order to decrease the time spent updating these structures, the updates were performed every 100 iterations or whenever the best known solution was improved. To provide the target/output values in the training data, the record solutions from \citep{optimizizer} were used. At each epoch, the algorithm provided a table of training data similar in structure to the one presented in Table \ref{table.logreg}. Using this table, at the end of each epoch we generated the maximum likelihood estimator for the logistic regression parameter \(\theta\) using the model (\ref{HF.compact}). Figure \ref{fig:regression} shows the average \(\theta\) values and the corresponding accuracy of the logistic regression model over 20 runs, along with the corresponding 95\% confidence intervals. 

The accuracy plots confirm that the best objective values \(D^1\) and \(D^0\) generated by the tabu search are relatively good predictors of the optimal solution values, with the average accuracy between 80\% and 95\%. Also, it is clear that the optimal regression parameter, as well as the model accuracy, tends to increase over time, until it reaches a saturation point when the algorithm is not able to find any improvements.   

\begin{figure}[!ht] \centering
\includegraphics[width=0.6\textwidth]{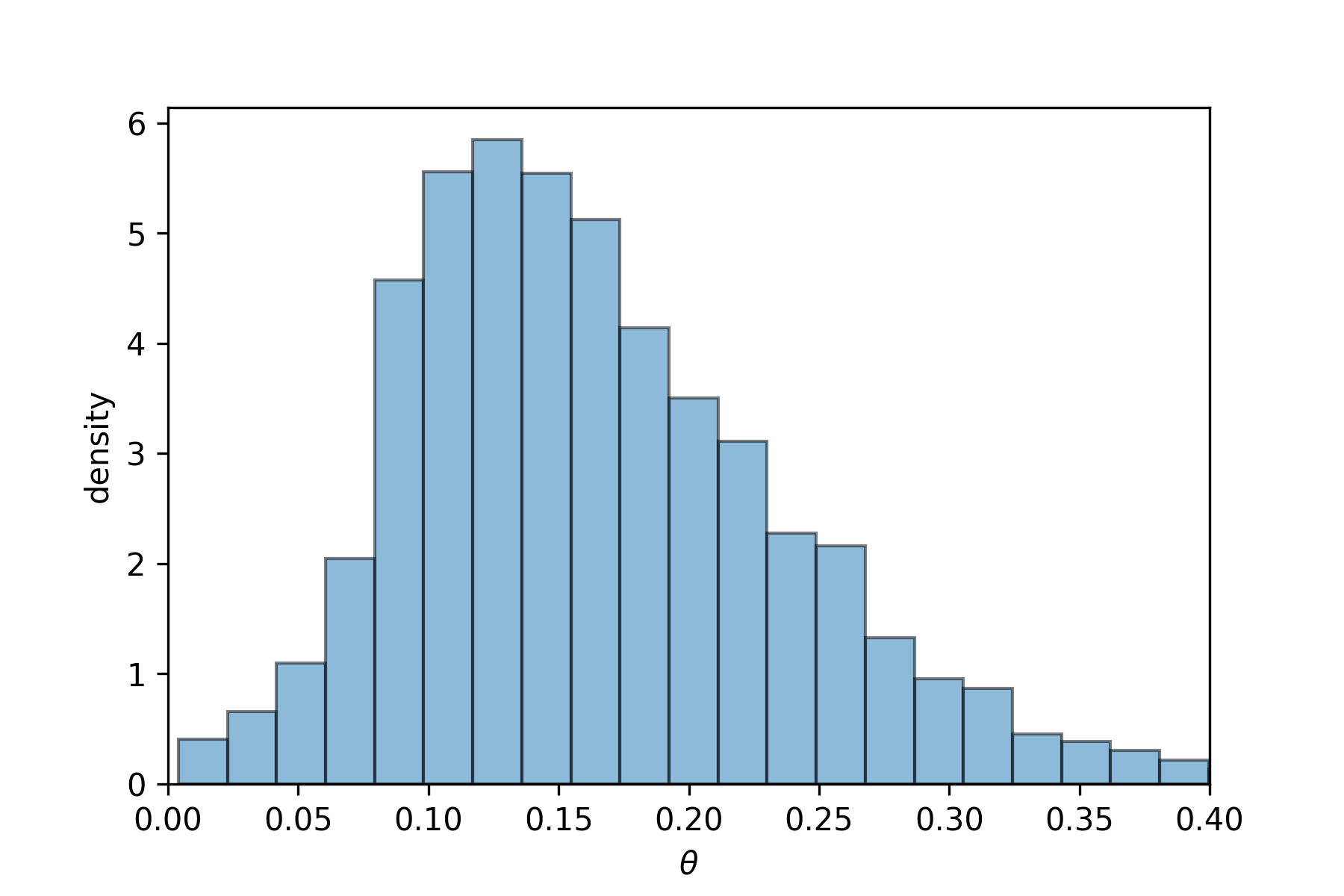}
     \caption{Distribution of the optimal regression parameter $\theta$ on instances ta11-ta50.}
     \label{fig:histogram.theta}
   \end{figure}

Figure \ref{fig:histogram.theta} shows the overall distribution of the optimal regression parameter \(\theta\) on all the instances.  Notice that the distribution of optimal \(\theta\) values is rather robust, and most of the  optimal \(\theta\) values belong to the interval [0,1]. It is important to emphasize that this range is specific for the tabu search algorithm and job shop problem instances ta11-ta50, and other algorithms may produce completely different ranges.  

\subsubsection{Backbone Size}
\label{sec:org5e2b10f}
The results of this section can be explored from the perspective of the big-valley and backbone results developed by \citet{Watson:2001,Streeter:2006}. The normalized backbone size \(\rho\) of the problem instance is defined as a proportion of binary variables that assume the same value in all optimal solutions. By considering random instances with the number of jobs \(N\) and the number of machines \(M\), it was shown that the normalized backbone size \(\rho\) depends on the ratio of \(N/M\). In the limit, for \(N/M \rightarrow 0\) the normalized backbone size converges to 1, while for \(N/M \rightarrow \infty\) it converges to 0. 

Assume that in the best schedule generated by the tabu search, given the backbone size \(\rho\), the average number of variables that share common values with the backbone variables is \(p_b \rho\). Given an optimal solution, let \(p_o (1-\rho)\) denote the average number of non-backbone variables in the best schedule that share common values with that optimal solution. Then the average logistic regression accuracy \(A\) can be expressed as 
$$
A = p_b \rho + p_o(1-\rho),
$$
where \(p_b\) and \(p_o\) can be viewed as measures of the algorithm performance. 

To explore this relationship, we conducted a test on the problem instance \texttt{ft10}  \citep{Beasley-1990-or-librar}. Using an arbitrary optimal solution, we performed 20 one-minute runs of the tabu algorithm, and the logistic regression reached 98\% accuracy on average. Since the algorithm found an optimal solution in each run, we have \(p_b=1\). Assuming \(p_o\geq 0.5\), the above expression for accuracy yields \(\rho \leq 0.96\), thus providing an upper bound on the normalized backbone size. Note that \citet{Streeter:2006} calculated the exact normalized backbone size of 95\% for this instance.

Next, we considered instances with a higher ratio of jobs to machines, and performed one-minute runs on five instances from \citep{Demirkol:1998}, dmu31-dmu35. In these problems the ratio of jobs to machines is equal to 3.33.  For each instance, we ran the tabu search 20 times and the algorithm found an optimal solution in each of the one-minute runs. Using an arbitrary optimal solution, we calculated that the average logistic regression accuracy over all five instances  was 0.61, which allow us to calculate an upper bound on the normalized backbone size, \(\rho\leq 0.23\).

\section{Guided Tabu Algorithm}
\label{sec:orgc60d28b}
\label{gta}

To tune the regression model for a specific problem instance, one needs to know an optimal or high-quality solution to that instance. When solving a new problem, the regression parameter has to be re-calibrated. Clearly, an optimal solution of the new problem instance is not available for such re-calibration, unless we solve the problem. Is it possible to use the logistic model from the previous section to boost the search process without knowing an optimal solution? 

Instead of finding the optimal regression parameter value, we generate a sequence of ``potential'' optimal values \(0=\theta_0\leq \theta_1\leq \ldots \leq \theta_K=1.0\), and scan through them one by one, generating model predictions. If the interval discretization is fine enough and a problem instance is similar to ta11-ta50, one of the \(\theta\) values may end up close to the optimal value, and the regression model will produce near-optimal predictions based on the available data.

Next, we outline the Guided Tabu Algorithm (GTA), which combines the standard tabu method and the logistic regression model described in the previous section. The proposed algorithm borrows some ideas from the Global Equilibrium Search method introduced by \citet{Shilo:1999}. 

\subsection{Algorithm Description}
\label{sec:orgef7b796}
\label{gta.desc}

\looseness=-1 Instead of using a single tabu tenure parameter, each component of a solution vector is assigned a separate tabu expiration parameter \(tabuExp(j)\), which provides the iteration at which the component's tabu status expires. These expiration times are updated whenever the component changes its value. By assigning large expiration times \(tabuExp(j)\) the algorithm attempts to preserve the current value of \(x_j\), while small expiration times \(tabuExp(j)\) will induce a faster rate of change for \(x_j\). 

The attribute-dependent tenures, where each solution attribute is assigned a separate tabu tenure value, has also been identified in earlier publications \citep{Glover:1993,Glover:1990a}. However, previous discussions of the attribute-dependent tenures mainly focused on the variability with respect to restrictive powers of different move attributes \citep{Glover:1993}, with an emphasis being placed on an idea that when using the same tabu tenure for all solution components, prohibition of certain solution attributes might have a stronger impact on the search process than prohibition of the others.

From the previous section, the reduced logistic regression model uses two values, \(D^1_j\) and \(D^0_j\), to predict the probability \(p_j\) that \(x^*_j=1\) in an optimal solution \(x^*\):

\begin{equation}
 p_j \equiv   \frac{1}{1+e^{\theta D^1_j(t) - \theta D^0_j(t)}} \approx P( x^*_{j}=1). \label{calc.prob}
\end{equation}

Notice that every variable with a tabu status can be in one of two states: \(x_j=1\) or \(x_j=0\). By setting distinct  tenure values for each of these states, we can assure that the ratio of overall time spend in each of them  is proportional to \(p_j\) and \(1-p_j\), respectively. This ratio can be achieved by using the following rule for updating tabu status expiration. Given a standard tenure duration \(\mathcal T\), if a local move changes the value of component \(j\) from \(x_j\) to \(x^{new}_j\) at iteration \(iter\), we can update the tabu expiration of the component \(j\) as
\begin{equation}\label{map.prob.to.tenure}
tabuExp(j) =   \left\{ 
\begin{array}{ll}
iter+  \mathcal T \frac{p_j}{\max[\epsilon,1-p_j]} & \text{if } (x^{new}_j =1 \text{ and } p_j\geq 0.5),\\
iter + \mathcal T  \frac{1-p_j}{\max[\epsilon,p_j]} & \text{if } ( x^{new}_j =0 \text{ and } p_j\leq 0.5),\\
      \mathcal T & \text{otherwise.} 
\end{array} 
\right.
\end{equation}

This mapping assures that if, after a local move, the new value of the variable \(x^{new}_j\) is not consistent with the model prediction \(p_j\), then it will be assigned some standard tenure duration \(\mathcal T\). Otherwise, the variable is assigned a larger tenure in proportion to the ratio between  \(p_j\) and \(1-p_j\) to promote a slower rate of change. The parameter \(\epsilon\) is a small constant that prevents division by zero (e.g., we use \(\epsilon = 10^{-4}\) in all computations). 

Procedure \ref{algo.gta} provides a description of the Guided Tabu Algorithm (GTA). Notice the similarities with the tabu search presented in Procedure \ref{algo.tabu}. Both algorithms operate for a number of epochs, each consisting of a fixed number of iterations. At each iteration, the algorithms scan through the neighborhood of the incumbent solution. The vector \(tabuExp\) keeps track of the tabu status of solution components. GTA uses additional data structures to keep track of the values \(D^1_j\) and \(D^0_j\) for each solution component. In order to decrease the computational overhead, the updates are performed every \(d\) iterations or whenever the best known solution is improved. After each update we calculate the probabilities according to (\ref{calc.prob}). These probabilities depend on the logistic regression parameter \(\theta\), which is defined as a function of the current iteration epoch:
$$
\theta=\theta_{min}\cdot\exp\left(\frac{\theta_{max}}{\theta_{min}}\cdot \frac{epoch-1}{nepochs-2}\right)
$$

The above expression assures that the values of \(\theta\) are log-uniformly distributed between \(\theta_{min}\) and \(\theta_{max}\). At iteration \(iter\), if the predictions provided by \(p_j\) are inconsistent with the new incumbent solution, the corresponding tabu expiration values are set based on the default tenure duration, \(tenure\). Otherwise, the algorithm assigns larger expiration values according to (\ref{map.prob.to.tenure}).

\begin{algorithm} \caption{Pseudocode of the Guided Tabu Algorithm. } \label{algo.gta}
\begin{algorithmic}[1]
\Procedure{GuidedTabuAlgorithm}{$T_{min}$, $T_{max}$, $nepochs$, $niters$, $\theta_{min}$, $\theta_{max}$}
    \State $tabuExp(j)=0$, $j=1,\ldots,|x^0|$;  \Comment{initialize tabu status expiration}  
    \State generate a random solution $x^{min}$; set $x^0 = x^{min}$, $\theta = 0$
    \For{$epoch=1$ to $nepochs$}
        \For{$iter=1$ to $niters$}     
            \State $tenure \leftarrow$ generate a random integer from $[T_{min},T_{max}]$
            \For{$x$ in $N(x^0)$}
               \State $expir(x) = iter$
               \For{$j$ in $\{j: x^0_j\neq x_j\}$}
                    \State $expir(x) = \max(tabuExp(j), expir(x))$                      
                \EndFor
                \If{$expir(x)>iter$}
                   \State TabuSet = TabuSet $\cup \ x$
                \EndIf
            \EndFor            
            \State NonTabuSet = $N(x^0)$ - TabuSet
            \If{NonTabuSet $\neq \emptyset$}
               \State $x^{new}= \arg \min \{f(x): x\in \text{NonTabuSet}\}$  \Comment{ best non-tabu solution}
            \Else
               \State $x^{new} = \arg \min\{ expir(x): x\in N(x^0)\}$ \Comment{ tabu solution with earliest expiration}
            \EndIf 
            \If{$iter \bmod d = 0 $ or $f(x^0)<f(x^{min})$} \label{updateD1D0}
               \State update vectors $D^1$, $D^0$ 
               \State calculate probabilities $p_j=1/\left(1+e^{\theta D^1_j(t) - \theta D^0_j(t)}\right)$,  $j=1,\ldots,|x^0|$;  
            \EndIf \label{endupdateD1D0}
            \For{$j$ in $\{j: x^{new}_j\neq x^0_j\}$} \Comment{only look at the components that have changed}
               \If{ ($x^{new}_j=1$ and $p_j\geq 0.5$) or ($x^{new}_j=0$ and $p_j\leq 0.5$)} \label{updateT}
                  \State $tabuExp(j)=iter+tenure\cdot \frac{\max(p_j,1-p_j)}{\max[\epsilon,\min(p_j,1-p_j)]}$                  
               \Else
                  \State $tabuExp(j)=iter+tenure$
               \EndIf \label{endupdateT}
            \EndFor            
            \State $x^0=x^{new}$
            \If{$f(x^0)<f(x^{min})$} 
               \State $x^{min}=x^0$
            \EndIf            
      \EndFor
      \State{$\theta = \theta_{min} \exp\left(\frac{\theta_{max}}{\theta_{min}}\cdot \frac{epoch-1}{nepochs-2}\right)$} \label{log.spacing}
    \EndFor
    \Return $x^{min}$
\EndProcedure
\end{algorithmic}    
\end{algorithm}

\subsection{Computational Study}
\label{sec:orge0f184b}

In this section, we compare tabu and guided tabu algorithms to  evaluate the impact of the logistic component. To provide intuitive and compact comparison, we introduce a new tool -- the probability dominance plots, which are based on the concept of probability dominance \citep{Wrather-1982-prob-domin}. 
These plots provide the answer to the following question: Given two algorithms, \(\mathcal A\) and \(\mathcal B\), a set of problem instances \(\mathcal C\), and the time threshold \(t\), what is the probability that the algorithm \(\mathcal A\) finds a better solution than the algorithm \(\mathcal B\) if each instance in \(\mathcal C\) is equally likely to be selected?

\subsubsection{Experimental Design}
\label{sec:org8a3f032}

\begin{longtable}{|c|c|c|}
\caption{\label{tab:orgb4f7ac1}
Notation for comparing algorithms.}
\\
\hline
Problem & Algorithm \(\mathcal A\) & Algorithm \(\mathcal B\)\\
\hline
\endfirsthead
\multicolumn{3}{l}{Continued from previous page} \\
\hline

Problem & Algorithm \(\mathcal A\) & Algorithm \(\mathcal B\) \\

\hline
\endhead
\hline\multicolumn{3}{r}{Continued on next page} \\
\endfoot
\endlastfoot
\hline
\(c_1\) & \(X^{\mathcal A}_{c_1}:= (x^1_1, x^1_2, \ldots,x^1_{n_1})\) & \(X^{\mathcal B}_{c_1}:= (y^1_1, y^1_2, \ldots,y^1_{m_1})\)\\
\(c_2\) & \(X^{\mathcal A}_{c_2}:= (x^2_1, x^2_2, \ldots,x^2_{n_2})\) & \(X^{\mathcal B}_{c_2}:= (y^2_1, y^2_2, \ldots,y^2_{m_2})\)\\
\(\cdots\) & \(\cdots\) & \(\cdots\)\\
\(c_k\) & \(X^{\mathcal A}_{c_k}:= (x^k_1, x^k_2, \ldots,x^k_{n_k})\) & \(X^{\mathcal B}_{c_k}:= (y^k_1, y^k_2, \ldots,y^k_{m_k})\)\\
\hline
\end{longtable}

For a pairwise comparison, we consider two algorithms \(\mathcal A\) and \(\mathcal B\) and a set of benchmark instances \(\mathcal C\). We are interested in estimating the probability of \(\mathcal A\) being better than \(\mathcal B\) with respect to one-dimensional performance measure when presented with a random instance from \(\mathcal C\) (each instance is equally likely to be selected).  To estimate this probability we repeatedly run \(\mathcal A\) and \(\mathcal B\) on each instance in \(\mathcal C\) and record corresponding performance measures for each run. Let \(X^{\mathcal A}_c\) denote a vector of performance measures obtained by repeatedly executing \(\mathcal A\) on a problem \(c\in \mathcal C\). The notation is clarified in Table \ref{tab:orgb4f7ac1}, where \(n_i\) corresponds to the total number of runs for \(\mathcal A\) on \(c_i\), and similarly \(m_i\) denotes number of runs for \(\mathcal B\) on \(c_i\).

When the goal is to minimize the performance measure, the probability that \(\mathcal A\) outperforms \(\mathcal B\) on a problem \(c\) can be estimated as 

$$
\displaystyle P_{\mathcal A< \mathcal B|c}=\frac{\displaystyle \sum_{x\in X^{\mathcal A}_c } \sum_{y\in X^{\mathcal B}_c} 1_{x< y} (x,y)}{|X^{\mathcal A}_c|\cdot |X^{\mathcal B}_c|}
$$
where \(1_{x< y}(x,y)\) is an indicator function

\begin{equation*}
1_{x< y}(x,y) = 
\begin{cases} 
1, & \text{if } x< y\\ 
0, & \text{otherwise.}
\end{cases}
\end{equation*}

The probability of \(\mathcal A\) outperforming \(\mathcal B\) on a random instance from \(\mathcal C\) is an average of \(P_{\mathcal A< \mathcal B|c}\) over all problems in \(\mathcal C\):
\begin{equation*}
P_{\mathcal A< \mathcal B|\mathcal C} = \frac{1}{|\mathcal C|}\sum_{c\in \mathcal C} P_{\mathcal A< \mathcal B|c}
\end{equation*}

The probability dominance plot shows the probabilities \(P_{\mathcal A< \mathcal B|\mathcal C}\), \(P_{\mathcal B< \mathcal A|\mathcal C}\) as a function of time, and the corresponding 95\% bootstrap confidence intervals \citep{Diciccio-1996-boots-confid-inter}. In the following computational experiments, the focus is on the best objectives found by the algorithms, but the same methodology can be applied to other measures of performance. The experimental data set and the code for generating the probability dominance plots are available on GitHub \citep{Shylo-2018-github-gtajob}.

\subsubsection{Computational Results}
\label{sec:org6fb4711}

\looseness=-1 To evaluate the impact of the logistic regression component on the performance, the tabu search from Section \ref{log.comp.sec} and the Guided Tabu Algorithm from Section \ref{gta.desc} were used to solve Taillard's instances ta11-ta50.
For both algorithms, we performed 20 runs on each problem instance, each run was limited to \(nepochs=200\) and each epoch consisted of \(niters=300,000\) iterations. The lower tenure range parameter \(T_{min}\) was set to 5 and \(T_{max}\) was set to 11 for both algorithms. The parameters \(\theta_{min}\) and \(\theta_{max}\) of GTA were set to 0.001 and 1.0, respectively. In GTA, the update frequency \(d\) was set to 100 iterations, and the parameter 
\(\epsilon\) in the tenure calculations was set to \(10^{-4}\).

\begin{figure}[!ht]
     \subfloat{%
       \includegraphics[width=0.5\textwidth]{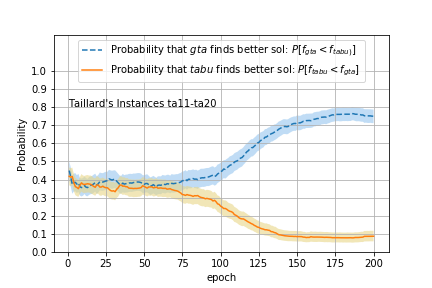}
     }
     \hfill
     \subfloat{%
       \includegraphics[width=0.5\textwidth]{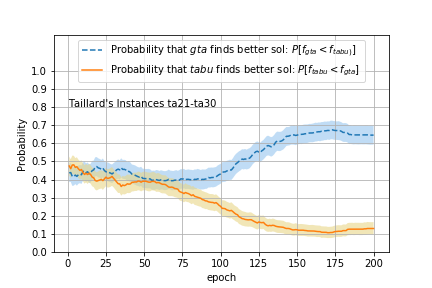}
     }\\[-0.9cm]
     \vfill
     \subfloat{%
       \includegraphics[width=0.5\textwidth]{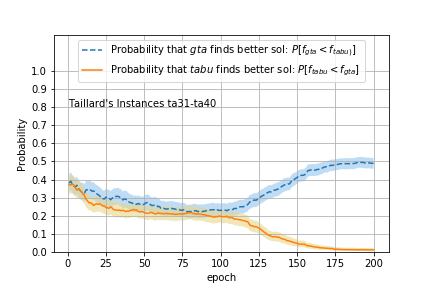}
     }
     \hfill
     \subfloat{%
       \includegraphics[width=0.5\textwidth]{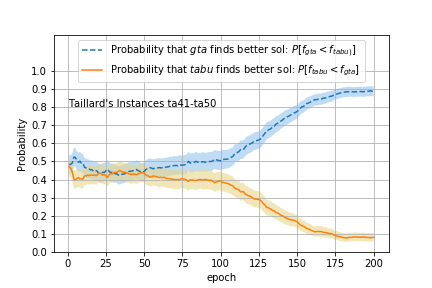}
     }
     \caption{Probability dominance plots comparing GTA and Tabu algorithm on the set of Taillard's benchmarks ta11-ta50 with respect to the \it{number of iterations}.}
     \label{fig:compta1150}
   \end{figure}

Figure \ref{fig:compta1150} shows the probability dominance plots with respect to the epoch numbers, using the best objective function values as the performance metric. As expected, in the initial epochs, when the logistic parameter \(\theta\) is small, GTA behaves identical to the standard tabu search, and the algorithms exhibit similar performance. However, as the time progresses, GTA is able to leverage information from the visited solutions to boost its performance. The average probability that the standard tabu search outperforms GTA after 200 epochs is less than 20\% on all problem instances.

\begin{figure}[!ht]
     \subfloat{%
       \includegraphics[width=0.5\textwidth]{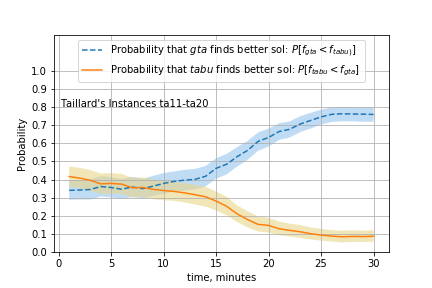}
     }
     \hfill
     \subfloat{%
       \includegraphics[width=0.5\textwidth]{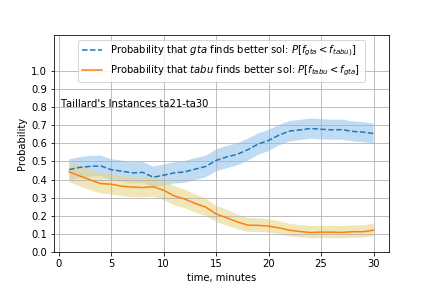}
     }\\[-0.9cm]
     \vfill
     \subfloat{%
       \includegraphics[width=0.5\textwidth]{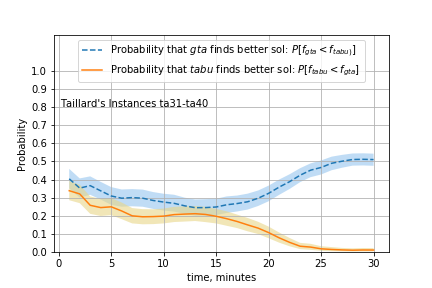}
     }
     \hfill
     \subfloat{%
       \includegraphics[width=0.5\textwidth]{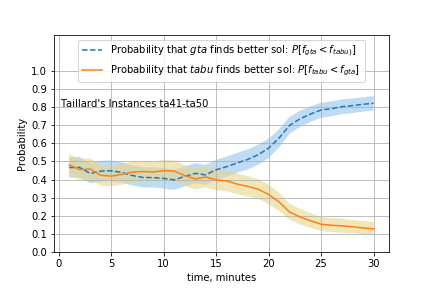}
     }
     \caption{Probability dominance plots comparing GTA and Tabu algorithm on the set of Taillard's benchmarks ta11-ta50 with respect to the \it{run times}.}
     \label{fig:compta1150-time}
   \end{figure}

Note that GTA spends more time in each epoch in order to evaluate \(D^1\) and \(D^0\). Though it is clear that the logistic component significantly improves computational performance in terms of solution quality, Figure \ref{fig:compta1150} does not provide much information about the actual run times. To check how the additional computations impact the time, we conducted an experiment, in which the duration of each run was fixed to 30 minutes and did not depend on the parameter \(nepochs\) (i.e., \(nepochs=\infty\)). All the parameters were identical to the previous experiment, except for the logistic parameter values. Instead of spacing them based on the current epoch number, line \ref{log.spacing} in Procedure \ref{algo.gta}, we updated the formula, so \(\theta\) is defined as a function of time. Given the threshold \(t^0\) (30 minutes in our computations), the parameter \(\theta\) at time \(t\) was calculated as \(\theta_{min} \exp\left(\frac{\theta_{max}}{\theta_{min}}\cdot \frac{t}{t^0}\right)\). Figure \ref{fig:compta1150-time} shows the corresponding probability dominance plots, which reveal that the tabu algorithm can be significantly improved by the logistic regression component.

\begin{figure}[!ht]
     \subfloat{%
       \includegraphics[width=0.5\textwidth]{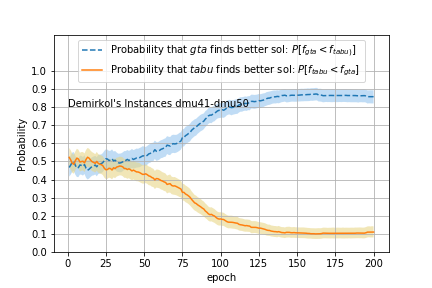}
     }
     \hfill
     \subfloat{%
       \includegraphics[width=0.5\textwidth]{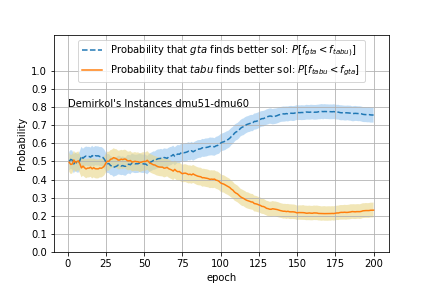}
     }
     \vfill
     \subfloat{%
       \includegraphics[width=0.5\textwidth]{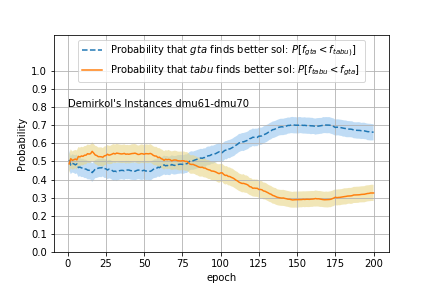}
     }
     \hfill
     \subfloat{%
       \includegraphics[width=0.5\textwidth]{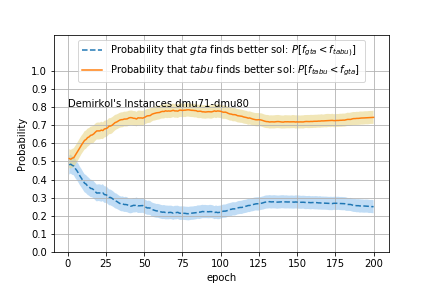}
     }
     \caption{Probability dominance plots comparing GTA and Tabu algorithm on the set of Demirkol's benchmarks dmu41-dmu80 with respect to the number of iterations.}
     \label{fig:compdmu4180}
   \end{figure}

\begin{figure}[!ht]\centering
       \includegraphics[width=0.5\textwidth]{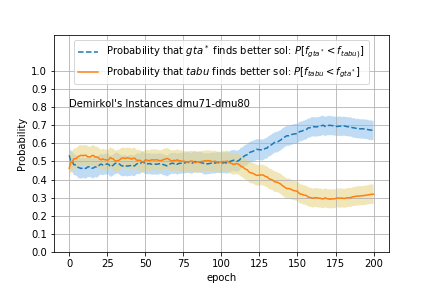}
     \caption{Probability dominance plots comparing GTA with $\theta_{min}=0.0001$ and Tabu algorithm on the set of Demirkol's benchmarks dmu71-dmu80 with respect to the number of iterations.}
     \label{fig:compdmu7180-delay}
   \end{figure}

The next computation experiment focused on the problems introduced by \citet*{Demirkol:1998}. This set contains 80 problems, denoted as dmu01-dmu80. As of now, the optimal solutions are known for 26 of these problems. The latest upper and lower bounds for these instances are publicly available \citep{optimizizer}. We limited our study to the instances dmu41-dmu80, which seem to be the hardest among those studied in the job shop scheduling literature. In these instances, the set of machines is split into two groups, and every job has to be processed by the machines in the first group before being processed by the second group.  This special routing structure seems to induce instances that are significantly harder than ta01-ta80. The number of binary variables in these problems range from 2850 to 24500 and the ratio between jobs and machines is between 1 and 3.3. Optimal solutions are not known for any of these problems. As of now, successful applications of the exact methods to these instances has never been reported in the literature. We used the same parameters as before, and the results of 20 GTA runs on each problem are presented in Figure \ref{fig:compdmu4180}. The results are similar to those presented for the Taillard's instances, with the exception of dmu71-dmu80, where the tabu search clearly outperforms GTA. Note that the probabilities diverge early on, which is different from all other instances. Clearly, the initial values of the logistic regression parameter are too large, causing premature fixation on early solutions with poor quality. To fix this issue, we changed \(\theta_{min}\) from 0.001 to 0.0001 and reran the experiment. Figure \ref{fig:compdmu7180-delay} shows the corresponding probability dominance plot. As observed on other instances, GTA traces the performance of the tabu search on early iterations and pulls away at the final epochs. 

\looseness=-1 As a byproduct, our experiments with GTA on the Taillard's and Demirkol's instances provided solutions (the new upper bounds are in the brackets) that improved existing upper bounds for the following problems: ta48 (1937), dmu42-dmu44 (3390, 3441, 3475), dmu48-dmu49 (3763, 3710), dmu51 (4156), dmu53-dmu55 (4390, 4362, 4270), dmu57-dmu65 (5655, 4708, 4619, 4739, 5172, 5251, 5323, 5240, 5190), dmu67 (5779), dmu68 (5765), dmu70 (5889), dmu72-dmu77 (6473, 6153, 6196, 6807, 6792), dmu79 (6952), dmu80 (6673) \citep{optimizizer}.

\section{Conclusions}
\label{sec:org71a0c2f}
\label{conclusions}

Our results show that the samples of visited solutions may contain valuable information that can be exploited via binary classification. The proposed framework is not limited to job shop scheduling and can be easily extended to any binary optimization problem. Many efficient solvers are good at finding high quality solutions, but do no attempt to prove optimality. In view of our results, it seems plausible that the solutions scanned by these solvers can be harnessed for optimality proofs. For example, the predictions of the presented regression model can guide branching and node selection decisions in branch and bound methods. 

Many optimization approaches rely on tabu method, but often utilize external methods for diversification and intensification of the search. For example, multi-start tabu strategies repeatedly launch tabu procedure from different initial solutions, which are generated using a subset of high-quality solutions \citep{Peng-2015-tabu-searc}. The algorithm proposed in this paper uses the native parameter of tabu search, tabu tenure, to provide intensification and diversification without complications common to similar extensions. The algorithm is easy to implement, has a small number of parameters and has shown excellent performance on the available job shop benchmark instances. The guided tabu algorithm can be easily adapted for any binary optimization problem.

\bibliographystyle{apalike}

\bibliography{overallliterature.bib}
\end{document}